# Supervised learning in Spiking Neural Networks with Limited Precision: SNN/LP

Evangelos Stromatias
School of Computer Science
The University of Manchester
Oxford Road, Manchester, United Kingdom
StromatE@cs.man.ac.uk

John S. Marsland
Dept. of Electrical Engineering and Electronics
University of Liverpool
Brownlow Hill, Liverpool, L69 3GJ, United Kingdom
marsland@liv.ac.uk

*Abstract*—A new supervised learning algorithm, SNN/LP, is proposed for Spiking Neural Networks. This novel algorithm uses limited precision for both synaptic weights and synaptic delays; 3 bits in each case. Also a genetic algorithm is used for the supervised training. The results are comparable or better than previously published work. The results are applicable to the realization of large scale hardware neural networks. One of the trained networks is implemented in programmable hardware.

*Keywords: Spiking neural networks; Supervised learning; Genetic algorithms; Limited synaptic precision; Temporal coding; Hardware implementation*

## I. Introduction

This paper reports on a new method (SNN/LP) for training spiking neural networks (SNN). There are a number of novel features in this work, most notably the use of limited precision (LP) for the synaptic weights and synaptic delays. The original motivation for using limited precision was to train hardware based SNNs. Limited precision gives a number of advantages for hardware neural networks in general, for example, digital implementations would have a lower bit count and consequentially a lower gate count, programmable implementations may have a reduced clock cycle count and analogue implementations would have a better noise immunity. The authors believe that a robust hardware solution that can be easily scaled would be one element of an overall system for the use of neural networks in real world applications. Although limited precision weights have been applied to traditional neural networks [1-3], this is the first attempt to apply limited precision to the weights and delays of an SNN synapse.

The second novel feature of this work is the use of a genetic algorithm (GA) for supervised learning in SNN. The binary encoding of the limited precision synapse, 6 bits per synapse for both weight and delay, allows the use of GAs. Previously Belatreche et al [4] have used an evolutionary strategy to train continuous valued SNN weights and delays, but limited precision and binary encoding lends itself to a GA approach. In this paper the GA training has been tested against two classic test bench problems; XOR and the Fisher Iris classification problem. The results compare very favorably with previous work reported by others. Indeed the performance of this new supervised training method, SNN/LP, would suggest that it is more generally applicable, not just a method of training hardware systems.

In the next section the background theory of SNN, limited synaptic precision and GAs is covered. In section 3, the results of training for the XOR problem are reported, for a number of different architectures comparable to previously reported work. In addition the results for a programmable hardware implementation are presented. In section 4, the Fisher Iris classification problem is addressed and results show that this new method, SNN/LP, gives a better performance than established methods such as SpikeProp, QuickProp and RProp. The final section draws some conclusions.

## II. Spiking Neural Networks and Limited Synaptic Precision

### A. The Spiking Neuron Model

Spiking Neural Networks have been referred to as the third generation of Artificial Neural Networks and it is known that they are more computational powerful than their predecessors [5] due to their capability to process information in a spatial-temporal way. In this work the Spike Response Model (SRM) was used to describe the behaviour of a spiking neuron. The SRM is an approximation of the Hodgkin-Huxley model [6] by a single-variable threshold model and it has been shown that it can predict 90% of the spikes correctly [7].

In SRM each time a neuron receives an input from a previous neuron its inner state, known as the membrane potential, changes. A neuron is said to fire a spike each time its membrane potential reaches a threshold value $\theta$ from below. Immediately after the firing of a spike the neuron goes through a phase of high hyperpolarization during which it is impossible to fire a second spike for some time, a time known as the refractory period. This undershoot is also known as the spike after potential [8]. The SRM used in this work is similar to [9] but slightly modified for a SNN with single synapse per connection. The membrane potential, $u$, of each neuron $j$ is described as:

$$u_j(t) = \rho(t - t_j^{(F)}) + \sum_{i=1}^{N_{i+1}} \sum_{g=1}^{G_i} w_{ji} \varepsilon(t - t_i^{(g)} - d_{ji}) \qquad (1)$$

Where $G_i$ are the total incoming spikes from the presynaptic neurons fired at time $t_i^{(g)}$, the $N_{l+1}$ are the presynaptic neurons to the $j^{th}$ neuron in layer l and finally the $w_{ji}$ and $d_{ji}$ represent the synaptic weights and delays. Function $\varepsilon$ calculates the unweighted postsynaptic potential (PSP) and $\rho$ is the refractoriness function which is responsible for the spike-after potential.

A neuron $j$ fires a spike at time $t_j$ when:

$$t = t_j^{(F)} \Leftrightarrow u_j(t) = \vartheta \quad and \quad \frac{du_i(t)}{dt} > 0 \qquad (2)$$

where $F$ is the total number of spikes that the neuron $j$ fired. The function $\varepsilon$ is expressed as:

$$\varepsilon(t_e) = \frac{t_e}{\tau} e^{1-\frac{t_e}{\tau}} H(t_e) \qquad (3)$$

and function $\rho$ is defined as:

$$\rho(t_p) = -4\vartheta e^{-\frac{t_p}{\tau_R}} H(t_p) \qquad (4)$$

where H is a Heaviside function and $\tau$, $\tau_R$ are time decaying constants.

### B. Limited Synaptic Precision

Limited weight precision has been investigated for traditional neural networks [1-3]. However there has no related work for SNN. This paper focuses on implementing SNN on hardware where the weights and delays can only be represented by a finite number of bits, resulting in a reduction of size, complexity and cost [1-3].

In this paper 6 bits were used to describe a synapse, Figure 1, and two different coding schemes have been investigated. Both of those schemes used integer values to describe the synaptic delays, expressed in milliseconds:

$$\{1, 2, 3, 4, 5, 6, 7, 8\}$$

The first coding scheme used a discretization of 0.5, that is a fractional value represented by one bit, for the weights:

$$\{-1.5, -1, -0.5, 0, 0.5, 1, 1.5, 2\}$$

While the second scheme used integer values for the weights:

$$\{-3, -2, -1, 0, 1, 2, 3, 4\}$$

### C. The Proposed Supervised Training Algorithm

The proposed supervised training algorithm is designed for limited precision feed-forward SNN with single synapse per connection that are allowed to emit multiple spikes and train both synaptic weights and delays.

The available supervised training algorithms such as SpikeProp [10] and its modifications QuickProp and RProp [11] allow their neurons to spike only once during the simulation time, thus not taking full advantage of the power of SNN. Furthermore, they can only use the time-to-first-spike coding scheme. This limitation is imposed in order to overcome the discontinuity of the threshold function.

Genetic Algorithms are a search and optimization method that is based on natural selection [12]. One of the advantages of the GA is that they can accommodate the discontinuity of the SRM. The algorithm used to train the SNN/LP is summarized in Table 1 and it is described more fully in the following sections.

In GA, each solution to a problem is encoded as a chromosome, an example of this is shown in Figure 1. After the encoding is established, a random population of chromosomes is generated, where the size of the population is set to be a multiple of the dimensionality of the problem. Subsequently, the objective function is calculated for each chromosome and the parents are selected based on the Baker's ranking scheme [13]. The mean-squared error is used as an objective function:

$$MSE = \frac{1}{N} \sum_{m=1}^{N} (t_j - t_j^d)^2 \qquad (5)$$

Where $N$ is the number of input patterns, $t_j$ is the actual output spike time and $t_j^d$ is the desired spike-time.

Uniform crossover is performed to the selected parents to produce new solutions. In Uniform Crossover, two individuals exchange genetic material based on a randomly generated mask. Mutation is then applied to the generated solutions, with a probability $p_m$, also known as the mutation rate and the randomly selected bits of the generated solutions are flipped.

**Table 1. The Proposed Genetic Algorithm.**

| | |
|---|---|
| Step 1: | Initialize a random starting population. |
| Step 2: | Check if convergence criteria have been established. |
| Step 3: | Evaluate the objective function F for all individuals of the population. |
| Step 4: | Rank all the individuals of the population based on their Objective values. |
| Step 5: | Perform parent selection based on the fitness probabilities of the individuals. |
| Step 6: | Apply crossover to selected parents with a crossover probability. |
| Step 7: | Apply mutation to the generated solutions based on a mutation rate. |
| Step 8: | Replace some parents with the offspring and create a new population. |
| Step 9: | If convergence criteria have not been met go to Step 2, else decode the best individual and pass it as the final answer. |

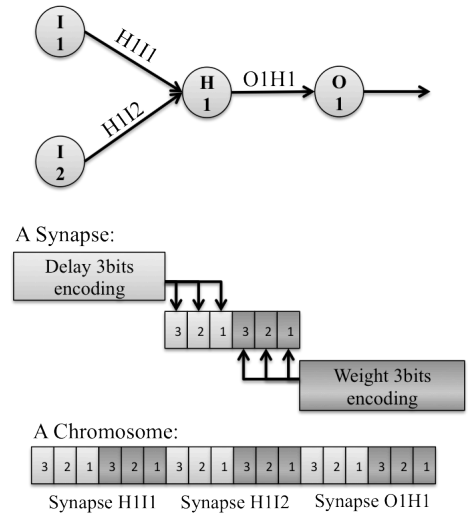

**Figure 1. The formation of a chromosome.**

The population is updated based on the generated solutions and the best individuals from the previous population, which pass unconditionally to the next one using the elitism operator. Finally, the GA is repeated until the convergence criteria is reached.

At this point, it should be noted that the proposed SNN/LP algorithm uses excitatory neurons and positive and negative weights like traditional ANN. However biological neural networks have neurons that are either excitatory, causing excitatory PSP, or inhibitory, causing inhibitory PSP, and the weights have positive values [8]. In this work neurons are not designated as inhibitory or excitatory before the training process, rather this is determined by the sign of the weight after the training has completed.

## III. THE XOR BENCHMARK

The XOR problem is a non linear classification problem that is often used as a first benchmark to supervised training algorithms in SNN, mainly because of its small dataset. Prior to any training however, the input and output spikes must be specified in terms of spike times. A simple method, proposed by [10], is used and it is shown in Table 2.

### A. Encoding of the XOR imputs and outputs

For the input patterns, logic zero is expressed as a spike occurring on the 1st millisecond, while logic one is expressed as a spike occurring on the 7th millisecond. For the output neuron, a spike fired at 17th millisecond indicates logic zero while a spike at the 10th millisecond indicates logic one. Also a third input is used, known as the reference neuron [10], which always fires a spike in the 1st millisecond.

**Table 2. The XOR problem encoded into spike-times (milliseconds).**

| Input reference neuron (R) | Input neuron 1 (I1) | Input neuron 2 (I2) | Output neuron (O1) |
|---|---|---|---|
| 1 | 1 | 1 | 17 |
| 1 | 1 | 7 | 10 |
| 1 | 7 | 1 | 10 |
| 1 | 7 | 7 | 17 |

### B. The network architecture and simulation settings

The SNN/LP algorithm is tested on the three different network architectures shown in Figure 2. The first network used 3 input neurons, 5 neurons in the hidden layer and 1 neuron in the output layer, Figure 2a. A similar network architecture was used in [10, 14, 15]; however, in this work, SNN/LP, only one synapse per connection is used (rather than multiple synapses with different delays) and only excitatory neurons with positive and negative values for the weights are used. Furthermore, the SNN/LP algorithm was able to train a network with fewer neurons. In [16] SpikeProp was tested on a smaller network than previous work. A 3-3-1 architecture was tested with 2 synapses per connection, each with different delays. In this work, a 3-2-1 architecture is investigated, Figure 2b.

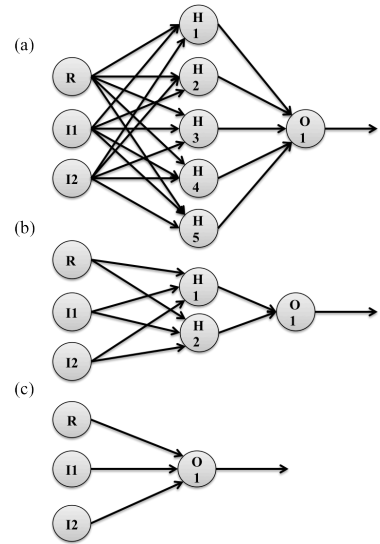

Figure 2. The three architectures that were considered.

Finally, as a last test, the training algorithm was able to solve the XOR problem without using a hidden layer. The same test was performed by [14] using real valued multiple delayed synapses, whereas here a single connection with limited precision per neuron is used. However, the threshold value of the SNN was increased to 3 and binary output coding was used for this topology. Logic zero results in no output spike while logic one is expressed as a spike in the 10th millisecond. Solving the XOR without a hidden layer is impossible for traditional neural networks [17].

All of the simulations were performed using Matlab for the two limited precision schemes (described in section II.B) and for two time steps, a 0.01 and 1. The SNN settings are shown in Table 3, while the parameters of the GA are given in Table 4. These parameters were chosen because they produced the best results after a series of tests with the same initial population.

**Table 3. The SNN Settings.**

| Simulation time | 50 ms |
|---|---|
| Tau | 3 |
| Tau R | 20 |
| Threshold | 1.5 |

**Table 4. The GA parameters.**

| Crossover Rate | 0.6 |
|---|---|
| Mutation Rate | 0.01 |
| Selective Pressure | 1.5 |
| Elitism Operator | 8 |
| Population size | 200 |

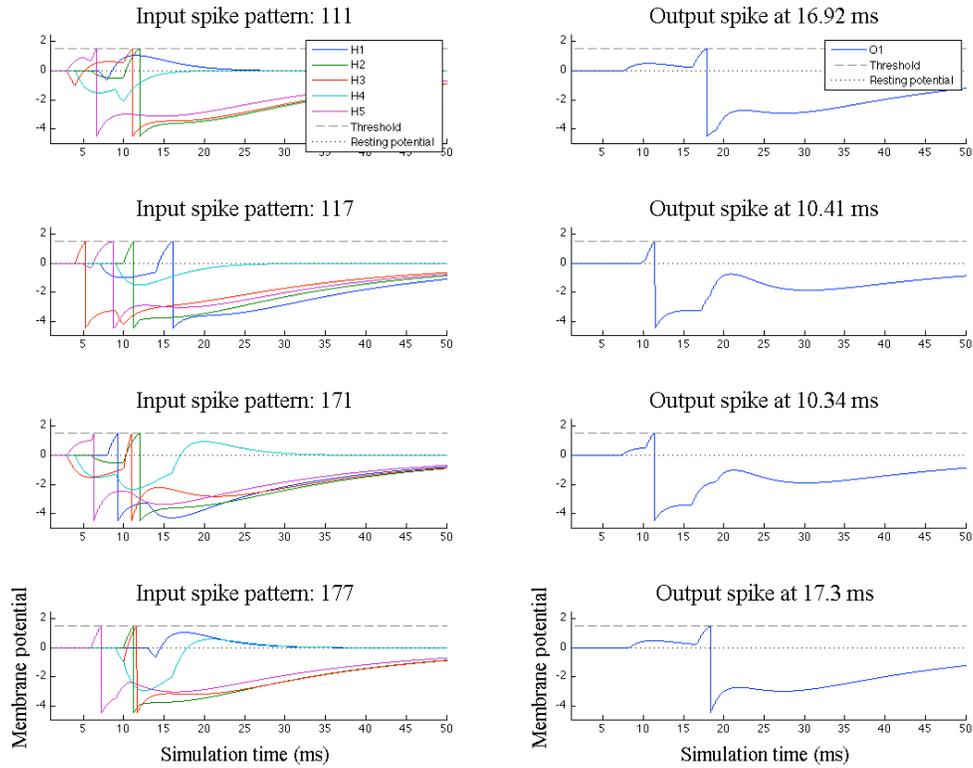

**Figure 3. The membrane potentials of the neurons in the hidden layer (left side) and in the output layer (right side), for a simulation time step 0.01.**

*C. Simulation results*

The membrane potentials of the neurons for all input patterns of the network architecture 3-5-1 can be seen in Figure 3. For more simulation plots and results, one should refer to [18].

The results are summarized in Table 5. The SNN/LP algorithm is able to converge to a much smaller MSE compared to SpikeProp, RProp and QuickProp [15]. When comparing these results to the aforementioned algorithms, one should bare in mind that this algorithm is designed for single synapse per connection whereas [10, 14, 15] use several sub-synapses, each with a different delay. This means that the SNN/LP algorithm needs fewer calculations for each neuron. Furthermore, this algorithm uses limited precision to train the synaptic weights and delays instead of real valued weights.

In [4] a supervised learning algorithm was developed based on Evolutionary Strategies (ES). Both synaptic weights and delays were trained with real numbers but the neurons were allowed to spike only once during the simulation time. In addition, their algorithm needed 450 generations to converge for the XOR (2-5-1 architecture), while the proposed algorithm converges in 40 generations for the one decimal binary weights, Figure 4, and in 21 generations for the integer weights, for the 3-5-1 architecture and for a 0.01 simulation time step [18].

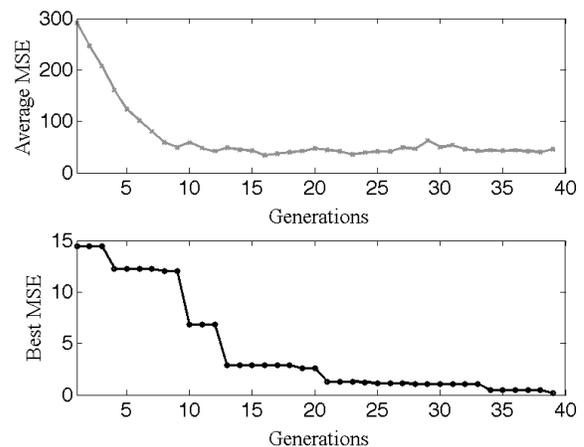

**Figure 4. Convergence of the Genetic Algorithm.**

**Table 5. Summary of the XOR training results in terms of MSE.**

| Network architectures | 1 binary decimal coding | | Integer coding | |
|---|---|---|---|---|
| | Time step 0.01 | Time step 1 | Time step 0.01 | Time step 1 |
| 3 – 5 – 1 | 0.09505 | 0 | 0.07135 | 0 |
| 3 – 2 – 1 | 0.2501 | 0.25 | 0.112625 | 0 |
| 3 – 1 | | 0.5 | | 1 |

## D. Hardware implementation of the single layer XOR

A hardware implementation of the single layer XOR was realized, in order to demonstrate that the trained synaptic weights and delays can be ported to a hardware system and produce results that are similar to the simulations. As a first step, a hardware simulation was performed using the Stateflow library from Simulink. Then the one-layered neuron was implemented on an ARM Cortex-M3 processor and the code was rewritten in the C programming language. Interrupts were used for the input spikes and, for simplicity, the four input patterns were saved in four buttons. In addition, the Matlab virtual simulation time was expressed as values of a software counter. The membrane potential and output spikes for each pattern of the integer weights coding scheme can be seen in Figure 5.

There is a small difference between the output spike-times of the Matlab simulations [18] and the hardware implementation and that is mainly due to the different source code implementation. In addition, as stated by [14] and as can be seen from Figure 5, for the input patterns {111} and {177}, the membrane potential approaches close to the threshold and, if the neuron is implemented in analogue, this might accidentally cause a spike due to noise in the signal.

used in SNN [10, 15, 20]. The centre $C_i$ and width $\sigma_i$ of each GRF was calculated by the following equations:

$$C_i = I_{min} + (\frac{2i-3}{2})(\frac{I_{max} - I_{min}}{m-2}) \quad (6)$$

$$\sigma_i = \frac{1}{\gamma}\frac{I_{max} - I_{min}}{m-2} \quad (7)$$

Where $\gamma$ is a constant number usually set to 1.5 [15, 20, 21]. The $I_{max}$ and $I_{min}$ values are the user defined maximum and minimum values of the input data and $m$ is the total number of the GRF neurons. Finally, a threshold value is used so that neurons that are below this threshold do not fire. The parameters of the GRF neurons can be seen in Table 6.

Table 6. The Gaussian Receptive Field neuron settings.

| m Gaussian Receptive Field neurons | 8 |
|---|---|
| Imin | 0 |
| Imax | 50 |
| γ | 1.5 |
| Threshold fire line | 0.1 |

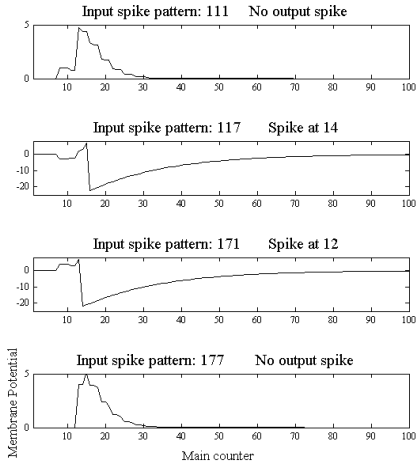

**Figure 5. The membrane potential for the hardware implementation of the one-layered XOR neuron.**

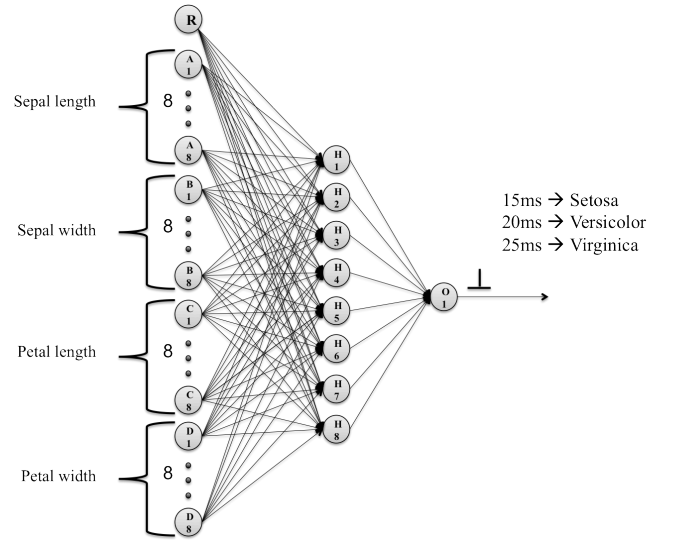

**Figure 6. The SNN architecture with the output encoding.**

## IV. THE FISHER IRIS CLASSIFICATION PROBLEM

In addition, the training algorithm was tested on the Fisher iris dataset [19], which consists of three classes of iris flowers: Setosa, Versicolor and Virginica with 50 samples each. Each of these classes has four input attributes: Sepal Length, Sepal Width, Petal Length and Petal Width. The Setosa class is linear separable to the other two classes, while the other two are not linear separable. A population of neurons [10] was used to convert the continuous real-valued features of the iris dataset into spike-times and this is described in the following section.

### A. Converting the Fisher iris attributes into spike times

In this work 8 Gaussian Receptive Field (GRF) neurons are used for each input iris feature. This method has been widely

### B. Network Architecture and Simulation Settings

The network architecture used can be seen in Figure 6. As mentioned in the previous section, 8 GRF neurons are used for each input feature, leading to a total of 33 input neurons, including a reference neuron. For the hidden layer, 8 neurons are used, same as [15]. Finally, the output layer had only one neuron firing a spike at 15ms to indicate the Setosa class, at 20ms to indicate the Versicolor class and at 25ms to indicate the Virginica class [15].

The simulation ran for 50 milliseconds, for a time step of 1 and for both limited precision schemes. The SNN settings were the same as Table 3 but on this occasion the threshold voltage for the one binary decimal weights was increased to 3 while the threshold voltage for the integer weights was increased to 6. The population size of the supervised training algorithm was increased to 600 and a maximum generation boundary was set to 600 in case the GA was not able to converge. The network was trained to a MSE of 0.25.

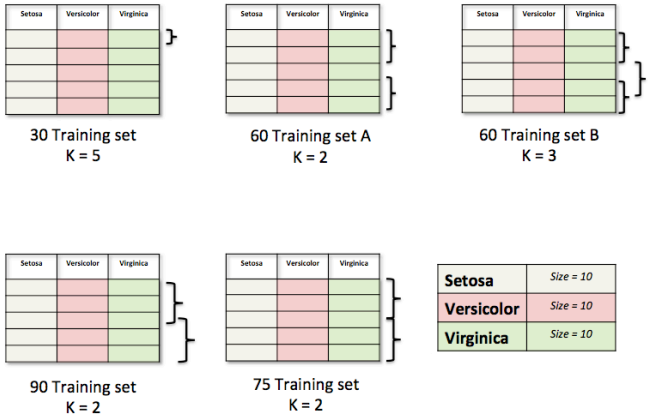

Figure 7. The 4 different training sets and the 5 K-Fold cross validation schemes that were used.

## C. Crossvalidation schemes and Results

The iris data set was split into the following training sets:
- A 30 training set, 10 input patterns from each class.
- A 60 training set, 20 input patterns from each class.
- A 75 training set, 25 input patterns from each class.
- A 90 training set, 30 input patterns from each class.

The K-Fold cross validation scheme was used for performance estimation. In K-Fold cross validation the dataset is split into K subsets, known as folds and each time one K subset is used for the training process while the remaining subsets are used for validation. Finally, the mean error is estimated as follows:

$$E = \frac{1}{K}\sum_{k=1}^{K} E_{val}(k) \quad (8)$$

The cross validation schemes are shown in Figure 7.

As in [15], in order for an output spike to be characterized as misclassified, it must be more than 2 milliseconds away from the desired spike-time. Finally, in Table 7, the SNN/LP algorithm is compared to SpikeProp, QuickProp and RProp [15].

Table 7: Classification accuracy between SpikeProp, QuickProp, RProp [15] and the SNN/LP algorithm.

| Training set | SpikeProp | QuickProp | RProp | SNN/LP 1 binary decimal | SNN/LP integer weights |
|---|---|---|---|---|---|
| 30 | 92.7% | 85.2% | 90.3% | 91.46% | 91.6% |
| 60 A | 91.9% | 91% | 94.8% | 96.89% | 95.56% |
| 60 B | 91.9% | 91% | 94.8% | 97% | 95.66% |
| 75 | 85.2% | 92.3% | 93.2% | 96% | 95% |
| 90 | 86.2% | 91.7% | 93.5% | 96.66% | 97% |

## V. CONCLUSION

A new supervised training algorithm, SNN/LP, has been developed in this work that is able to train fully connected feed forward SNNs with single synapse per connection and with limited synaptic precision. Moreover, the algorithm allows each neuron to fire multiple spikes thus making it possible to use coding schemes other than time-to-first-spike.

The efficiency of the algorithm, however, lies in the fact that both synaptic weights and delays are trained. So, even though 6 bits were used to describe a synapse, 3 for the synaptic delays and 3 for the weights, the proposed algorithm was able to achieve very low MSEs for the XOR problem with less neurons than previously reported and classification accuracies up to 97% for the Fisher iris problem [18].

The authors believe that these results will be very useful for future implementations of SNNs on hardware as it will be possible to include more neurons in the same design without sacrificing the size, cost and the total accuracy.

ACKNOWLEDGMENT

This work formed part of the MSc project of the first author [18], completed in partial fulfillment of the MSc(Eng) degree from the University of Liverpool. The authors are grateful for the use of the High Throughput Computing service, CONDOR, provided by the University of Liverpool [22].